\documentclass{article}
\usepackage{fancyhdr}
\usepackage{spconf,amsmath,graphicx}
\usepackage{enumerate}
\usepackage{booktabs}
\usepackage{tabularx}
\usepackage{url}
\usepackage[hidelinks]{hyperref}
\usepackage{xurl}          
\fancypagestyle{firstpage}{
  \fancyhf{}
  
}

\title{Modeling AI-Driven Production and Competitiveness: A Multi-Agent Economic Simulation of China and the United States}
\name{Yuxinyue Qian, Jun Liu}
\address{Beijing University of Posts and Telecommunications\\
liujun@bupt.edu.cn}
\begin{document}
%
\maketitle
\thispagestyle{firstpage}

\begin{abstract}
With the rapid development of artificial intelligence (AI) technology, socio-economic systems are entering a new stage of "human–AI co-creation." Building upon a previously established multi-level intelligent agent economic model, this paper conducts simulation-based comparisons of macroeconomic output evolution in China and the United States under different mechanisms—AI collaboration, network effects, and AI autonomous production. The results show that: (1) when AI functions as an independent productive entity, the overall growth rate of social output far exceeds that of traditional human-labor-based models; (2) China demonstrates clear potential for acceleration in both the expansion of intelligent agent populations and the pace of technological catch-up, offering the possibility of achieving technological convergence or even partial surpassing. This study provides a systematic, model-based analytical framework for understanding AI-driven production system transformation and shifts in international competitiveness, as well as quantitative insights for relevant policy formulation.

\end{abstract}
\begin{keywords}
Socio-economy; Intelligent Agents; AI Collaboration; Autonomous Production Mechanism; China–U.S. Comparison
\end{keywords}
\section{Introduction}
\label{sec:intro}

Since the beginning of the 21st century, the rapid evolution of generative artificial intelligence (AI) and autonomous intelligent agents (AI agents) has profoundly reshaped the operating mechanisms of socioeconomic systems. Overall, the United States maintains a significant lead in core model development and capital investment. Public data indicate that in 2023, the U.S. launched 109 representative foundation models—far exceeding China’s 20 models \cite{StanfordHAI2024_RnD}. In the same year, private-sector AI investment in the U.S. reached USD 67.2 billion, compared with China’s USD 7.8 billion \cite{StanfordHAI2024_Economy}. Furthermore, the United States possesses a far more extensive computing infrastructure, with 5,381 data centers compared to approximately 449 in China \cite{Lu2025_DataCenters}, providing a more stable foundation for large-scale model training and inference. In contrast, China has promoted the rapid diffusion of AI applications through centralized implementation and deployment mechanisms. The proportion of enterprises that regularly use generative AI (83\%) surpasses that of the United States (65\%), though China still slightly lags behind in "deep implementation"—that is, end-to-end automation and process reengineering (19\% vs. 24\%)\cite{Fisher2025_AIDominance}. This contrast underscores a crucial conclusion: China leads in the breadth of AI application, while the United States holds the advantage in depth of technology and ecosystem maturity.

Against this backdrop, there is an urgent need to systematically assess the differing impacts of intelligent agents on socioeconomic output across the two countries, taking into account their distinct national contexts. Existing research has preliminarily confirmed that the introduction of AI agents enhances both micro-level productivity and macro-level welfare. However, quantitative analyses comparing China and the United States remain limited.

On one hand, China’s vast labor force—approximately 770 million people \cite{NBS2019}—and substantial capital accumulation \cite{PWT2024_ChinaCapital} provide the scale and data availability necessary for large-scale agent collaboration and autonomous production. On the other hand, the United States’ first-mover advantages in computing power, algorithms, and business models \cite{TrendForce2024_AIServers},\cite{StanfordHAI2025_TechPerformance} may enable it to enter a "virtuous cycle" of agent adoption → network effects → productivity growth earlier. Consequently, these structural asymmetries are likely to shape how AI contributes to aggregate output in each country. Key questions arise: To what extent can AI, as a tool for collaborative enhancement, improve human labor efficiency in each context? How much incremental value can it generate as an autonomous productive entity? And will network externalities among agents manifest differently between China and the United States? Clarifying these issues is vital for formulating targeted AI development policies and unlocking the potential of the intelligent economy.

This paper proposes a comparative analysis based on five progressive intelligent-agent economic model frameworks to evaluate the output performance of China and the United States following AI-agent integration. The structure of the paper is as follows: Section II reviews the existing literature; Section III outlines the model construction and explains parameter settings; Section IV presents simulation results for both countries under different models, emphasizing how differences in population size, total resources, AI efficiency and penetration, and network effects shape the outcomes; Section V discusses the underlying mechanisms and analyzes China’s potential for accelerated improvement; Section VI offers policy recommendations and future outlooks.

Through this framework, the study aims to uncover: (1) the extent to which intelligent-agent collaboration enhances output in each country and which structural parameters drive this effect; (2) the relative strength of mechanisms such as collaborative enhancement, autonomous production, and network externalities; (3) which modes of AI-agent integration best suit the national conditions and development stages of China and the U.S.; and (4) which policy levers can be adjusted to further unleash AI’s potential.

\section{Related work}
\label{sec:format}

The evolution of artificial intelligence is commonly understood as being driven by the synergistic advancement of three fundamental pillars—computing power, algorithms, and data \cite{MIT2025_AIProgress}. This triadic framework has become a core analytical lens in both academic and policy discussions of AI development. Against this backdrop, the present study examines the divergent trajectories of China and the United States across four interrelated dimensions: computing infrastructure, models and algorithms, scientific and knowledge output, and application and commercialization. On this basis, it further identifies key research gaps in the existing literature.

\noindent\textbf{Computing Power. }A growing body of studies and industry reports suggests that the pace and ceiling of AI iteration are largely constrained by the availability of computing resources \cite{Tomasev2025_VirtualAgents}. The United States has long maintained a leading position in high-end GPU manufacturing \cite{TrendForce2024_AIServers}, the scale of data centers \cite{Lu2025_DataCenters}, and network infrastructure \cite{TeleGeography2024_Network}, providing robust hardware support and enabling large-scale model training. In contrast, China continues to face bottlenecks in advanced chip fabrication and semiconductor processes \cite{Reuters2025_ChinaChips},\cite{Ezell2024_Semiconductors}, limiting the overall maturity of its hardware ecosystem.
However, China’s position in the field of intelligent computing power is not entirely subordinate. According to a white paper by the China Academy of Information and Communications Technology (CAICT), China accounts for approximately 39\% of global intelligent computing capacity, compared with 31\% for the United States \cite{CAICT2024_ComputeIndex}. This indicates that China already possesses notable advantages in the total volume and engineering deployment of AI-oriented computational resources, with substantial potential for optimization and coordinated scheduling.

\noindent\textbf{Models and Algorithms. }The performance gap between Chinese and American AI models is narrowing rapidly. Based on the MMLU (Massive Multitask Language Understanding) benchmark, the performance difference between the top models of the two countries stood at 17.5 percentage points in 2023, but had decreased to just 0.3\% by early 2024 \cite{StanfordHAI2025_TechPerformance}. This trend demonstrates China’s capacity for swift convergence through advances in algorithm design, training strategies, and engineering optimization.

\noindent\textbf{Scientific Research and Knowledge Output. }At the level of research and knowledge production, the structural divergence between China and the United States can be characterized as a contrast between quantity and breadth versus quality and frontier leadership. China ranks first globally in both patent applications and academic publication volume, with the number of generative AI–related patent filings over the past decade reaching six times that of the United States, reflecting vast research capacity and talent reserves. Yet, in terms of citation impact and research influence, the United States maintains a clear advantage. For instance, among the 100 most-cited papers in AI, the U.S. contributed 50, while China contributed 34, with U.S. papers averaging about 15\% higher citation rates \cite{StanfordHAI2025_RnD}. This pattern suggests that China’s research output remains more application-oriented and quantity-driven, whereas the United States continues to dominate in foundational theory and paradigm innovation.

\noindent\textbf{Application and Commercialization. }The divergence between the two countries is even more pronounced in AI application and commercialization pathways. The U.S. model centers on the consumerization and SaaS-based enterprise deployment of generative AI. By 2024, approximately 68\% of U.S. enterprises had adopted generative AI—up from 33\% just one year earlier \cite{StanfordHAI2025_Economy}—and the U.S. AI software and services sector accounts for roughly 60\% of the global market \cite{IDC2024_Spending}.
China, by contrast, emphasizes industrial integration, process modularization, and scenario-driven customization. In the manufacturing sector, for example, China’s annual installations of industrial robots account for about 52\% of the global total \cite{StanfordHAI2025_Economy}, underscoring its distinctive advantage in industrial applications. Nevertheless, Chinese AI firms still lag behind in overall profitability. U.S. companies such as OpenAI generated approximately USD 3.7 billion in revenue in 2024, with Anthropic reaching USD 0.9 billion \cite{Edward2025_Revenue}, reflecting a mature commercial ecosystem. Chinese firms, while exhibiting large-scale deployment potential and extensive application coverage, remain smaller in revenue scale and often depend on specific vertical scenarios, with business models still under development.

\noindent\textbf{Synthesis. }Taken together, these dynamics reveal a multidimensional and evolving landscape rather than a static gap. The United States continues to lead in financial strength, advanced chip technology, high-impact research quality, and the maturity of its commercial ecosystem. China, meanwhile, demonstrates robust potential through its rapid progress in model performance, the scale of research output, and the expanding domestic application market.
Despite a wealth of research on the effects of AI on productivity, employment, and industrial structure—typically focusing on a single country or mechanism—there remains a lack of systematic, model-based comparison of the structural and dynamic differences between China and the United States. To address this gap, this study develops a unified modeling framework to explicitly compare the two nations across key structural parameters, assess the relative strength of different AI-driven mechanisms, and identify actionable policy levers to guide future development pathways for the intelligent economy.

\section{Modeling Methodology}
\label{sec:maintitle}

To explore the impact of generative AI agents on socio-economic collaboration, we construct five progressively extended models.  
Model 1 describes a purely human collaboration scenario;  
Model 2 introduces AI agents as collaborators;  
Model 3 further incorporates network effects among AI agents;  
Model 4 assumes that AI becomes an independent productive entity sharing resources with humans;  
Model 5 combines features of Models 3 and 4, considering both AI independent production and network synergy.  
Each model reflects varying degrees of human-AI collaboration through different mechanisms.  
The following are the core mathematical formulations (the derivation process follows prior work \cite{Qian2025_AIAgentModel}).

\subsection{Model 1: Pure Human Collaboration Model}
The total social output depends on population and resource inputs:
\begin{equation}
Y(t) = \phi_0 N^{\alpha} R^{1 - \alpha},
\end{equation}
where \( Y(t) \) is the total social output, \( \phi_0 \) is the baseline efficiency without AI involvement,  
\( N \) is the human population, \( R \) is the total amount of resources, and \( \alpha \) is the output elasticity of human resources.

\subsection{Model 2: Collaboration Model with AI Agents}
In this model, total resources \( R \) are divided between humans and AI agents:
\begin{equation}
R = R_H + R_A,
\end{equation}
where \( R_H \) is the portion of resources used by humans, and \( R_A \) is that used by AI agents.

The evolution of AI capability is characterized by an S-shaped function:
\begin{equation}
s_t = \frac{1}{1 + e^{-k(t - t_0)}},
\end{equation}
where \( k \) is the growth rate and \( t_0 \) is the inflection point of capability improvement.  
The value of \( s_t \in [0,1] \) indicates the relative progress of AI technology from its initial to mature stage.

The total output is given by:
\begin{equation}
Y(t) = \phi_0 N^{\alpha} R_H^{1 - \alpha} 
\left[ 1 + \gamma \left( \frac{R_A}{R_H} \right)^{\beta} (1 + \delta s(t))^{\beta} \right],
\end{equation}
where \( \gamma \) is the baseline coefficient of AI enhancement efficiency and \( \beta \) is the elasticity of AI enhancement.

\subsection{Model 3: AI Collaboration Model with Network Effects}
Building on Model 2, we incorporate the network externality brought by the number of AI agents \( A(t) \):
\begin{equation}
\begin{split}
  Y(t) &= \phi_0 N^{\alpha} R_H^{1-\alpha} 
   \left[1 + \gamma \left(\frac{R_A}{R_H}\right)^{\beta} (1 + \delta s(t))^{\beta}\right] \\
   &\quad \times \left(1 + \eta \left(\frac{A(t)}{N}\right)^2\right).
\end{split}
\end{equation}
where \( \eta > 0 \) represents the maximum amplification magnitude of the network effect.

\subsection{Model 4: AI as an Independent Productive Entity}
Let the share of resources used by AI be defined as:
\begin{equation}
\omega = \frac{R_A}{R}.
\end{equation}

Then, the total output function is:
\begin{equation}
\begin{split}
  Y(t) &= \phi_H N^{\alpha} \left[(1 - \omega)R\right]^{1-\alpha} \\
      & + \phi_A A(t)^{\alpha} \left[\omega R (1 + \delta s(t))\right]^{1-\alpha}.
\end{split}
\end{equation}
where \( \phi_H \) and \( \phi_A \) denote the baseline efficiencies of human labor and AI production, respectively.

\subsection{Model 5: Integrated Model with Independent AI Production and Network Effects}
Finally, combining Model 4 with network effects, the total output becomes:
\begin{equation}
\begin{split}
  Y(t) &= \phi_H N^{\alpha} \left[(1 - \omega)R\right]^{1-\alpha} \\
       &+ \phi_A A(t)^{\alpha} \left[\omega R (1 + \delta s(t))\right]^{1-\alpha} 
         \left(1 + \eta \left(\frac{A(t)}{N}\right)^2\right).
\end{split}
\end{equation}

\section{Simulation Analysis}

Based on the five models described above, we design a computational simulation experiment to quantitatively analyze the impact of AI agent participation in collaboration on the total social output in China and the United States.

\subsection{Parameter Settings}

\begin{enumerate}[(1)]
    \item \textbf{Total Population ($N$)} \\
    In 2019, the employed population in China was approximately 770 million \cite{NBS2019}, 
    while that in the United States was about 159 million \cite{PWT2024_EmploymentUS}. 
    Thus, the population parameters are set as:
    \begin{equation}
        N_{cn} = 7.7 \times 10^8, \quad N_{us} = 1.59 \times 10^8.
    \end{equation}

    \item \textbf{Total Resources ($R$)} \\
    The total resource input is represented by the total capital stock.     
    In 2010, China's total capital stock reached USD 39.3 trillion, 
    and the U.S. total capital stock was USD 61 trillion.
    In 2019, China's total capital stock reached USD 99.6 trillion \cite{PWT2024_ChinaCapital}, 
    and the U.S. total capital stock was USD 69.1 trillion \cite{PWT2024_CapitalUS}. 
    Therefore:
    \begin{equation}
        R_{2010,cn} = 3.93 \times 10^{13}, \quad R_{2010,us} = 6.1 \times 10^{13}.
    \end{equation}
    \begin{equation}
        R_{2019,cn} = 9.96 \times 10^{13}, \quad R_{2019,us} = 6.91 \times 10^{13}.
    \end{equation}

    \item \textbf{Capital Allocation Structure ($R_H$ and $R_A$)} \\
    According to China Daily (2023), the adoption rate of generative AI among Chinese enterprises has reached 15\% \cite{ChinaDaily2023_AIIndustry}. 
    We therefore assume that 15\% of the total capital is allocated to AI agents and 85\% to human labor:
    \begin{equation}
        R_A = 0.15 R, \quad R_H = 0.85 R.
    \end{equation}

    \item \textbf{Labor Output Elasticity ($\alpha$)} \\
    Based on national statistics, the labor share in 2019 was 0.58625 for China \cite{PWT2024_LabShareCN}
    and 0.59709 for the United States \cite{PWT2024_LabShareUS}. Hence:
    \begin{equation}
        \alpha_{cn} = 0.58625, \quad \alpha_{us} = 0.59709.
    \end{equation}

    \item \textbf{AI Enhancement Elasticity ($\beta$)} \\
    By 2040, generative AI is projected to increase labor productivity by 0.1\% to 0.6\% per year \cite{McKinsey2023_GenAIPotential}. 
    To capture the strength of this trend, the elasticity parameter is set as:
    \begin{equation}
        \beta = 0.35.
    \end{equation}

 \item \textbf{AI Capability Scaling Factor ($\delta$)} \\
    Industry reports indicate that generative AI can significantly enhance efficiency across various business domains. 
    For example, customer operations may improve by approximately 30\%--45\%, marketing by 5\%--15\%, 
    sales by 3\%--5\%, and software engineering by 20\%--45\% \cite{McKinsey2023_GenAIPotential}. 
    These findings suggest that AI exhibits strong potential for productivity enhancement in practical applications. 

    To reasonably estimate the efficiency gain of AI in the model while accounting for inter-industry variation 
    and real-world technological penetration, the AI capability scaling factor is set to:
    \begin{equation}
        \delta = 0.20.
    \end{equation}
    This value implies that when AI technology reaches maturity, 
    its resource utilization efficiency can increase by approximately 20\% over the baseline level.

    \item \textbf{Baseline Coefficient of AI Enhancement Efficiency ($\gamma$)} \\
    According to Google's experimental studies, 
    software engineers using AI-assisted coding tools such as \emph{Copilot} 
    improved their task completion speed by 55.8\% \cite{Peng2023_GitHubCopilot}. 
    Based on this empirical evidence, the baseline coefficient is set as:
    \begin{equation}
        \gamma = 0.55.
    \end{equation}

    \item \textbf{AI Capability Level ($s(t)$)} \\
    The evolution of AI technological capability, denoted as $s(t)$, 
    follows a logistic function. 
    The parameters of its growth rate and inflection point are defined as follows:

    \begin{enumerate}[(a)]
        \item \textit{Growth Rate ($k$)} \\
        For China, the rate of technological diffusion is calibrated using 
        the logistic fitting of China's internet penetration between 2001 and 2011, 
        yielding $k_{cn} = 0.38$ \cite{Wu2013_InternetDiffusion}. 
        In contrast, the organizational adoption rate of generative AI in the United States 
        is significantly lower than that of China (approximately 65\% vs. 83\%) \cite{Fisher2025_AIDominance}. 
        Consequently, the U.S. exhibits slower diffusion and a lower effective productivity gain, 
        and thus we set:
        \begin{equation}
            k_{us} = 0.30.
        \end{equation}

        \item \textit{Inflection Point ($t_0$)} \\
        In China, \textit{DeepSeek} was officially released on January 20, 2025, 
        and rapidly gained traction in the AI domain. 
        The platform reached 100 million users within seven days, 
        and by February 2025, the number of generative AI users in China had reached 250 million. 
        Hence, the technological development inflection point for China is set as:
        \begin{equation}
            t_{0, cn} = 5.
        \end{equation}

        In the United States, a key milestone for generative AI advancement 
        was the launch of \textit{ChatGPT} in late 2022, 
        which markedly improved accessibility and usability for text-based tasks. 
        The subsequent release of \textit{GPT-4} in March 2023 represented another major leap in AI capability. 
        Therefore, we define the U.S. inflection point as:
        \begin{equation}
            t_{0, us} = 3.
        \end{equation}
    \end{enumerate}

    \item \textbf{Network Effect Coefficient ($\eta$)} \\
    The maximum amplification coefficient of network effects $\eta$ 
    is derived from empirical research on the LinkedIn platform. 
    The regression results indicate that under different income dimensions, 
    the coefficient corresponding to Metcalfe’s law ranges between 0.064 and 0.077 \cite{Schin2023_LinkedInEffect}, 
    implying that when user penetration approaches 100\%, 
    network effects can generate approximately 6\%–8\% additional value. 
    Based on these empirical findings, the benchmark setting is:
    \begin{equation}
        \eta = 0.07.
    \end{equation}

    \item \textbf{Number of AI Agents ($A(t)$)} \\
    Following the model assumption of linear growth, 
    the number of AI agents over time is defined as:
    \begin{equation}
        A(t) = A_0 + g t,
    \end{equation}
    where $A_0$ denotes the initial number of AI agents, 
    and $g$ represents the annual growth rate of AI agents.

    \item \textbf{Initial Number of AI Agents ($A_0$)} \\
    As of June 2024, the number of generative AI users in China reached 230 million \cite{Xinhua2024_GenAIUsers}. 
    According to a survey, about 65\% indicated that their organizations 
    regularly used generative AI in at least one business function \cite{McKinsey2024_StateOfAI}. 
    Thus, the estimated number of users actively employing generative AI for collaborative work is:
    \begin{equation}
        2.3 \times 10^8 \times 65\% = 1.495 \times 10^8.
    \end{equation}
    Therefore, the initial AI agent quantity in China is set as:
    \begin{equation}
        A_{0, cn} = 1.495 \times 10^8.
    \end{equation}

    According to the National Bureau of Economic Research (NBER), 
    approximately 28\% of U.S. employees have used generative AI in their work \cite{NBER2024_GenAIAdoption}. 
    Hence, the initial AI agent quantity in the United States is:
    \begin{equation}
        A_{0, us} = 1.59 \times 10^8 \times 0.28 = 4.45 \times 10^7.
    \end{equation}

     \item \textbf{Annual Growth of AI Agents ($g$)} \\
    According to public reports, the annual increase in the number of domestic AI agents exceeds 10 million, 
    which is approximately 85 times the number of new applications added annually to the Apple App Store. 
    This figure provides a realistic reference for determining the growth rate parameter $g$. 

    Considering that not all AI agents can be effectively transformed into productive collaborative units, 
    this study sets the feasible range of $g$ between 3 million and 10 million agents per year. 
    In the baseline scenario, we define:
    \begin{equation}
        g_{cn} = 5 \times 10^6.
    \end{equation}

    Assuming that the United States maintains the same \emph{relative growth rate} as China, we have:
    \begin{equation}
        r_{cn} = \frac{g_{cn}}{A_{0, cn}} 
        \approx \frac{5.0 \times 10^6}{1.495 \times 10^8} 
        \approx 3.34\%.
    \end{equation}

    Thus, for the United States:
    \begin{equation}
        g_{us} = r_{cn} \times A_{0, us} 
        \approx 0.0334 \times 4.45 \times 10^7 
        \approx 1.486 \times 10^6.
    \end{equation}

    \item \textbf{AI Resource Share ($\omega$)} \\
    The parameter $\omega$ measures the proportion of AI-related capital 
    (including algorithms, computational power, and data infrastructure) 
    within the total national production resources, 
    which encompass human capital, physical capital, and AI capital.

    According to the World Economic Forum (WEF), 
    approximately 22\% of current work tasks are primarily performed by machines or algorithms \cite{Josh2025_AIJobs}, 
    providing an upper bound for the substitution effect of AI on human labor. 
    However, when AI is viewed as a capital input rather than direct labor replacement, 
    its overall share in total production resources remains relatively low.

    Moreover, China exhibits a higher adoption rate of generative AI than the United States 
    (approximately 83\% vs. 65\%), 
    while the U.S. leads in application maturity (24\% vs. 19\%) \cite{Fisher2025_AIDominance}. 
    This implies that China's AI usage is more widespread but at an earlier stage of automation, 
    whereas the U.S. demonstrates deeper and more mature integration. 

    Therefore, we set:
    \begin{equation}
        \omega_{cn} = 0.05, \quad \omega_{us} = 0.15,
    \end{equation}
    reflecting China’s broader yet less automated AI adoption 
    and the United States’ higher technological maturity.

    \item \textbf{Baseline Efficiencies ($\phi_0$, $\phi_H$, and $\phi_A$)} \\
    We employ a calibration method to determine the baseline efficiency levels 
    for each country under conditions with and without AI participation. 

    In 2010, China’s GDP was USD 6.19 trillion, and the United States’ GDP was USD 15.05 trillion.  
    By 2019, China’s GDP had risen to USD 14.58 trillion \cite{wikipediaHistoricalGDPChina2025}, 
    while that of the United States reached USD 21.54 trillion \cite{wikipediaEconomicStatisticsUnited2025}.  
    Therefore, we define:
    \begin{align}
        Y_{2010, \mathrm{cn}} &= 6.19 \times 10^{12}, &
        Y_{2010, \mathrm{us}} &= 15.05 \times 10^{12}, \notag\\
        Y_{2019, \mathrm{cn}} &= 14.58 \times 10^{12}, &
        Y_{2019, \mathrm{us}} &= 21.54 \times 10^{12}.
    \end{align}

    Using the 2010 values of $Y_{2010}$, $N$, and $R$, 
    we can solve for the baseline human efficiency:
    \begin{equation}
        \phi_0 = \phi_H = \frac{Y_{2010}}{N^{\alpha} R_{2010}^{1 - \alpha}}.
    \end{equation}

    This yields approximately:
    \begin{equation}
        \phi_{H, cn} \approx 90, \quad \phi_{H, us} \approx 532.
    \end{equation}

    Given these baseline human efficiencies, 
    we can decompose the total 2019 output into human and AI contributions 
    (assuming $\omega = 0.1$, $s_t = 0.5$, $\delta = 0.2$, and $A(t) = 10^8$):
    \begin{equation}
        Y_H = \phi_H N^{\alpha} \left[(1 - \omega) R_{2019} \right]^{1 - \alpha}, 
        \quad Y_A = Y_{2019} - Y_H.
    \end{equation}

    The baseline AI efficiency can then be back-calculated as:
    \begin{equation}
        \phi_A = \frac{Y_A}{A^{\alpha} \left[\omega R_{2019} (1 + \delta s)\right]^{1 - \alpha}}.
    \end{equation}

    The resulting AI baseline efficiencies are:
    \begin{equation}
        \phi_{A, cn} \approx 483, \quad \phi_{A, us} \approx 688.
    \end{equation}    
\end{enumerate}

\subsection{Baseline Scenario: Pure Human Collaboration between China and the United States}

\begin{figure}[htbp]
    \centering
    \includegraphics[width=0.45\textwidth]{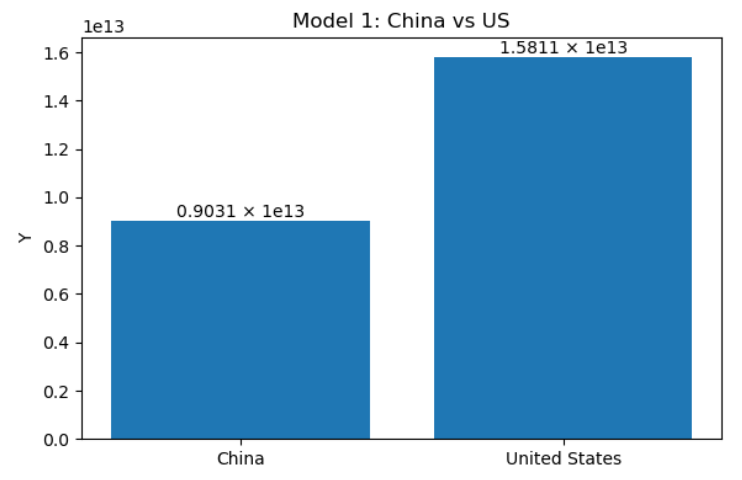}
    \caption{Model 1: Comparison of outputs between China and the United States 
    under pure human collaboration scenario}
    \label{fig:model1_cn_us}
\end{figure}

In the baseline model without AI participation, the total social output of China and the United States 
depends entirely on their respective endowments of human and material resources. 
As shown in Figure~\ref{fig:model1_cn_us}, based on the calibrated parameters, 
the total social outputs of China and the United States are 
$9.031 \times 10^{12}$ and $15.911 \times 10^{12}$, respectively, 
with an output ratio of approximately $1:1.75$. 
This result indicates that although the United States does not possess an advantage 
in total resource quantity, its substantially higher per-capita productivity 
constitutes the fundamental reason for its significantly greater aggregate output.

\subsection{Collaboration Model with AI Agents}

Model~2 introduces AI agents as collaborators working alongside humans in the production process. 
In this model, the inclusion of AI consumes a portion of the total resources, 
but it also enhances human labor efficiency, thereby increasing overall output. 

\begin{figure}[htbp]
    \centering
    \includegraphics[width=0.45\textwidth]{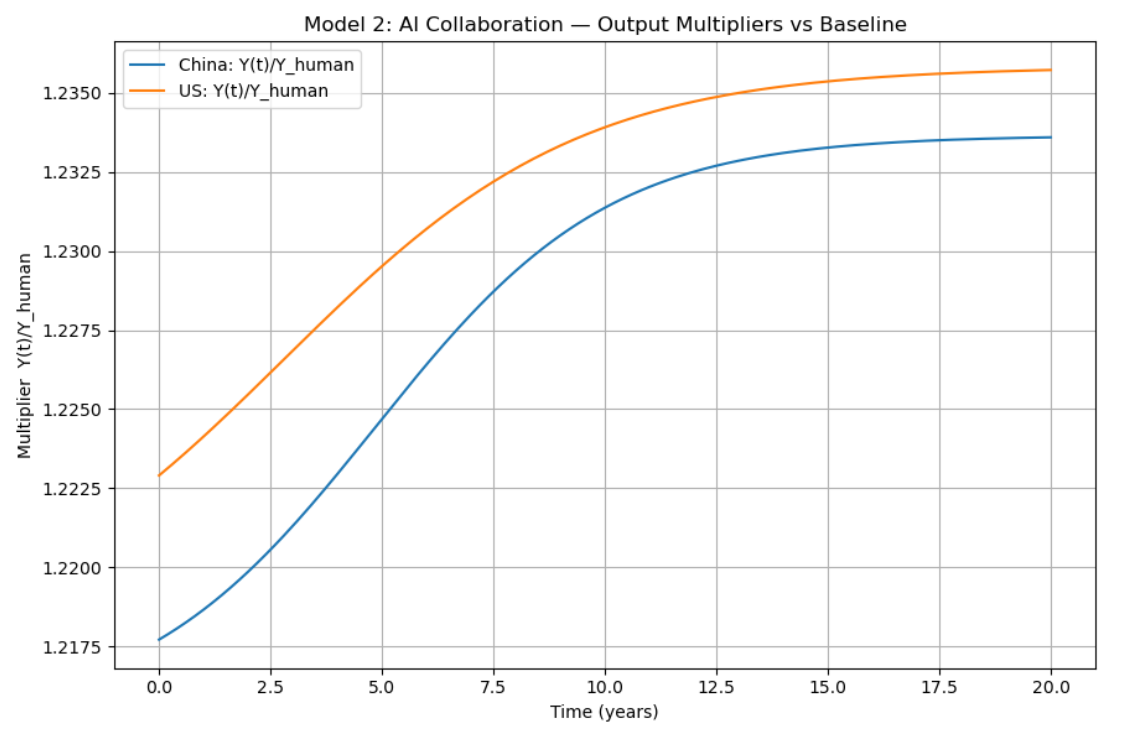}
    \caption{Model~2: Comparison of output multipliers between China and the United States 
    under the AI collaboration scenario}
    \label{fig:model2_cn_us}
\end{figure}

Figure~\ref{fig:model2_cn_us} illustrates the time evolution of the output multiplier 
of China and the United States relative to the baseline model (Model~1). 
The results show that the output multipliers of both countries increase over time 
and eventually converge to a steady state, 
confirming that AI-driven productivity gains exhibit a developmental phase 
followed by a saturation period. 

However, the U.S. growth curve (orange) consistently remains above that of China (blue) 
throughout the entire time horizon. 
This persistent gap suggests that the United States, 
owing to its initial advantages in AI technology deployment 
(such as more advanced computational infrastructure) 
and higher collaboration efficiency, 
is able to translate AI-induced productivity gains 
into economic output earlier and more effectively.

\subsection{AI Collaboration Model with Network Effects}

Building upon the AI collaboration framework, 
Model~3 further incorporates the network externalities among AI agents. 
The core mechanism of this model is that as the number of AI agents grows linearly, 
their penetration rate $p$ increases over time, 
which in turn amplifies total output through a network effect multiplier 
defined as $\Theta(p) = 1 + \eta \cdot p^2$. 

\begin{figure}[htbp]
    \centering
    \includegraphics[width=0.45\textwidth]{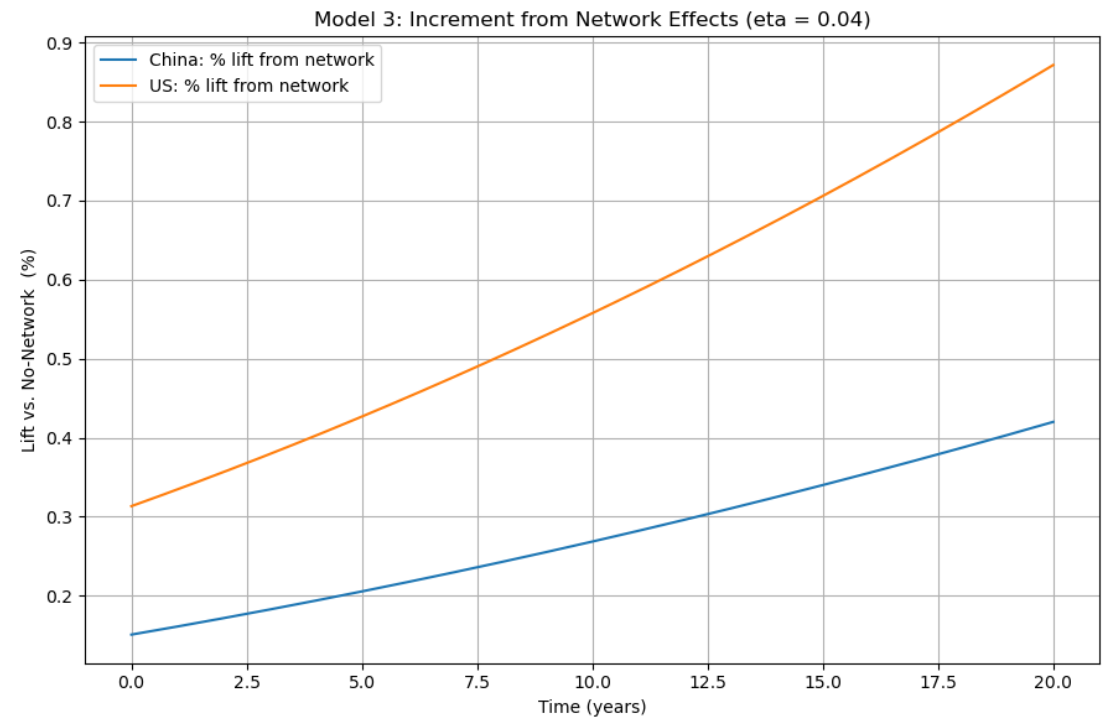}
    \caption{Model~3: Comparison of marginal enhancement effects 
    on total social output between China and the United States 
    under network-effect AI collaboration}
    \label{fig:model3_cn_us}
\end{figure}

We simulate both national scenarios under a network intensity parameter of $\eta = 0.04$. 
Figure~\ref{fig:model3_cn_us} illustrates the marginal enhancement effect of network externalities 
on total social output for China (blue line) and the United States (orange line). 
As shown, under the same network effect setting, 
the total output in both countries rises steadily over time, 
indicating that the growth of AI agents and the increase in penetration rate 
continuously strengthen productivity gains derived from network externalities. 

However, compared with the baseline scenario, 
the enhancement effect in the United States remains significantly higher than that of China 
throughout the entire simulation period, 
and the gap between the two continues to widen: 
the initial improvement is approximately 0.31\% for the United States 
versus 0.15\% for China, 
which expands to about 0.87\% versus 0.42\% by the end of the period. 
This widening divergence is fundamentally driven by 
differences in the growth trajectories of AI penetration ($p$) between the two countries, 
suggesting that a higher initial penetration rate or a faster diffusion speed 
can generate a more pronounced competitive advantage under network effects.

\subsection{AI as an Independent Productive Entity}

Model~4 explores another scenario in which AI is endowed with independent production capabilities, 
allowing it to share resources with humans and generate output autonomously. 
This model corresponds to a highly automated economic structure, 
where AI no longer merely assists humans but can independently accomplish a portion of productive tasks. 

\begin{figure}[htbp]
    \centering
    \includegraphics[width=0.45\textwidth]{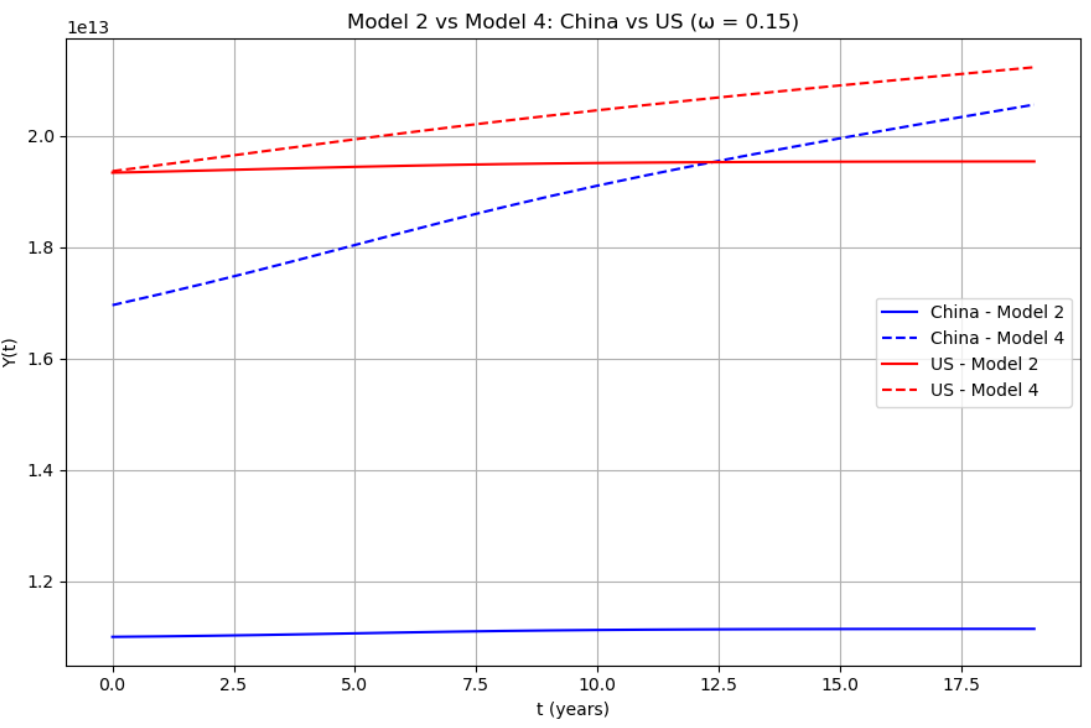}
    \caption{Comparison of total social output between China and the United States 
    under Model~2 (AI collaboration) and Model~4 (AI independent production) 
    with identical AI resource share ($\omega = 0.15$)}
    \label{fig:model4_cn_us}
\end{figure}

Figure~\ref{fig:model4_cn_us} compares the total social output trajectories of China and the United States 
under the same AI resource share ($\omega = 0.15$) 
for Model~2 (AI collaboration) and Model~4 (AI independent production). 
The results indicate that the absolute output level of the United States 
remains consistently higher than that of China in both models. 
However, a noteworthy observation is that in Model~4, 
China’s output curve exhibits a significantly steeper slope, 
demonstrating a strong upward trend. 
This suggests that in a scenario where AI functions as an independent productive agent, 
China could leverage the rapid expansion and diffusion of its AI agents 
to offset its initial efficiency disadvantage, 
thereby gradually narrowing—and potentially surpassing—the output gap with the United States 
over the medium to long term. 

Further comparison between the two models reveals that 
the productivity gains in Model~2 are relatively limited: 
the output curves of both countries eventually plateau, 
as growth primarily depends on AI’s enhancement of human labor efficiency. 
This dynamic enables the United States, with its higher baseline productivity 
and more mature enterprise-level AI application ecosystem, 
to maintain its dominant position. 
In contrast, Model~4 allows for sustained marginal output growth 
through the endogenous expansion of both the number and capabilities of AI agents. 
It is precisely through this mechanism that China has the potential 
to progressively close the output gap with the United States.

\subsection{Comprehensive Model: AI as an Independent Productive Entity with Network Effects}

\begin{figure}[htbp]
    \centering
    \includegraphics[width=0.45\textwidth]{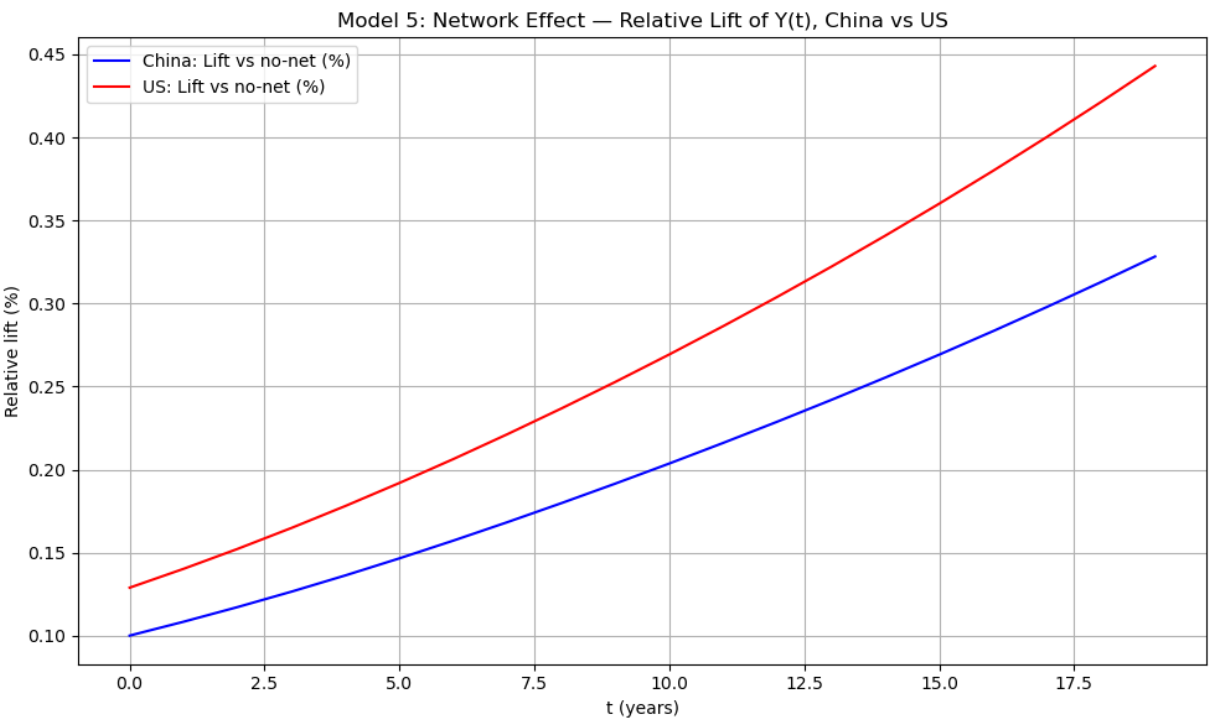}
    \caption{Model~5: Comparison of network-effect-induced output enhancement 
    between China and the United States under the comprehensive model}
    \label{fig:model5_cn_us}
\end{figure}

Model~5 integrates both independent AI production and network externalities. 
The simulation results (Figure~\ref{fig:model5_cn_us}) compare 
the relative enhancement of total social output for China and the United States 
under this comprehensive framework. 

The results show that the relative improvement in the United States 
remains consistently higher than that in China throughout the entire simulation period. 
Specifically, at the initial time point, the network effect contributes 
approximately a $0.13\%$ increase in social output for the United States, 
compared to $0.10\%$ for China. 
After twenty years, the improvement in the United States rises to about $0.44\%$, 
while that in China reaches $0.33\%$. 
This persistent gap primarily stems from differences in the initial number and growth rate 
of AI agents, as well as variations in the share of AI-related resources. 

Although the initial scale of AI agents in the United States is smaller than that of China, 
its higher AI resource share
and more mature application ecosystem accelerate the marginal diffusion of network effects, 
thereby generating stronger output gains and reinforcing its productivity advantage.

\section{Discussion}

Significant differences exist between the development trajectories of artificial intelligence in China and the United States. 
China exhibits a clear trend of ``accelerated advancement'' in several key parameters, 
providing potential for technological catch-up and even partial overtaking. 
Based on the framework of Model~4, 
this section focuses on identifying which parameters demonstrate acceleration potential for China.

\subsection{Number of AI Agents}

Since there is currently no standardized international metric for quantifying software-based AI agents, 
this study adopts the number of \textit{physical AI agents}---that is, the number of industrial robots installed annually---as a proxy variable. 
An industrial robot is defined as 
``an automatically controlled, reprogrammable, multipurpose manipulator programmable in three or more axes, 
which may be either fixed in place or mobile, and is applied in industrial automation scenarios.'' 

\begin{figure}[htbp]
    \centering
    \includegraphics[width=0.45\textwidth]{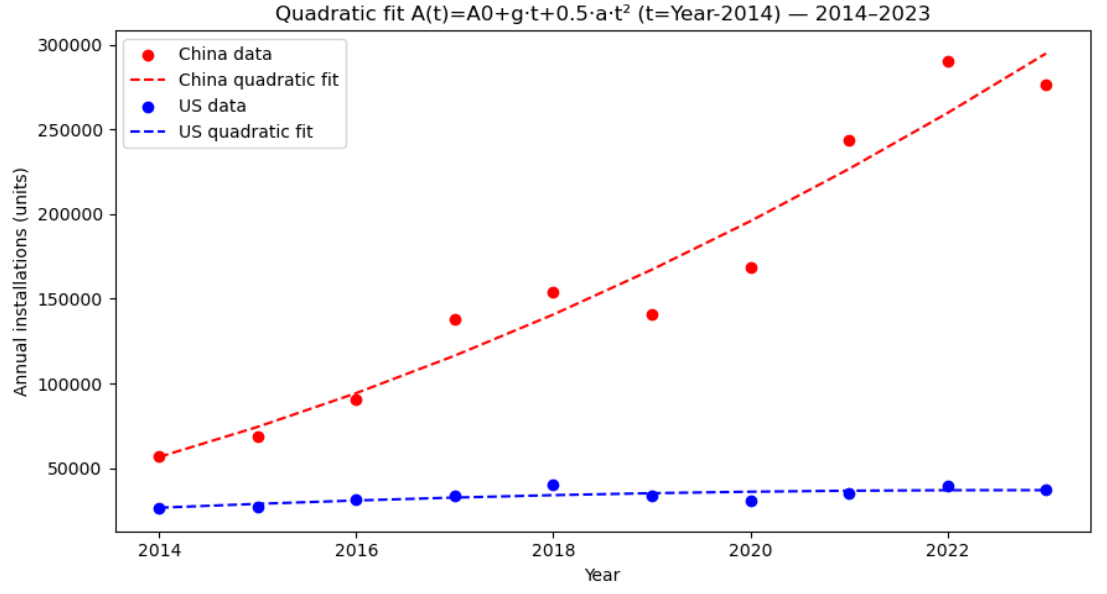}
    \caption{Fitted growth curves of AI agent quantities in China and the United States (2014--2023)}
    \label{fig:fit_agents_cn_us}
\end{figure}

Using data from 2014 to 2023, 
we fit a quadratic growth model to the number of AI agents in China and the United States, 
expressed as:
\begin{equation}
A(t) = A_0 + g t + \tfrac{1}{2} a t^2.
\end{equation}
The fitted results are shown in Figure~\ref{fig:fit_agents_cn_us}. 
For China, the estimated model is:
\begin{equation}
A_{\text{cn}}(t) = 56{,}612 + 16{,}674t + 1{,}088t^2,
\end{equation}
where the positive quadratic coefficient ($a = 1{,}088$) 
indicates a pronounced acceleration effect in the growth trend. 
For the United States, the fitted model is:
\begin{equation}
A_{\text{us}}(t) = 26{,}648 + 2{,}416t - 140t^2,
\end{equation}
where the negative quadratic coefficient ($a = -140$) 
suggests a deceleration trend, with growth momentum gradually weakening over time.

These results demonstrate that, in terms of AI agent quantity, 
China not only leads in total scale but also exhibits a sustained capacity for accelerated growth. 
In contrast, the United States experienced a slowdown after reaching a stage peak around 2018. 
Based on this evidence, 
we assume that with effective policy intervention, 
China’s annual growth in AI agents can maintain a relatively high rate. 

To reflect the gradual increase in the annual number of new AI agents, 
the growth rate of China’s AI agents, $g_{\text{cn}}(t)$, 
is set as an increasing function of time in the simulation. 
The baseline initial value is set to $g_{\text{cn}}(0) = 5 \times 10^6$, 
consistent with the previous benchmark. 
Considering the strong acceleration in industrial robot installations between 2014 and 2023 
and the continued expansion of generative AI applications, computational power, 
and data infrastructure thereafter, 
we introduce a robust acceleration path under ceteris paribus conditions: 
$g_{\text{cn}}(t)$ increases by 20\% relative to the baseline by the end of the 20-year simulation period, 
that is, $g_{\text{cn}}(19) = 6 \times 10^6$. 
The baseline scenario follows a linear acceleration form:
\begin{equation}
g_{\text{CN}}(t) = 5 \times 10^6 + 5 \times 10^4 t.
\end{equation}

\begin{figure}[htbp]
    \centering
    \includegraphics[width=0.45\textwidth]{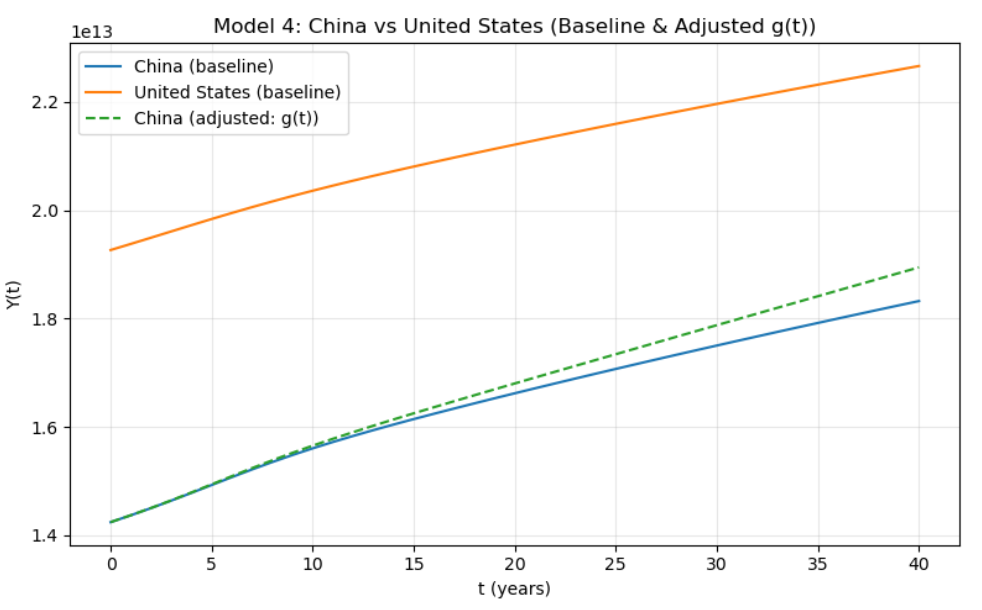}
    \caption{Comparison of total output under adjusted AI agent growth rate ($g$)}
    \label{fig:model7_agent_growth}
\end{figure}

The simulation results are shown in Figure~\ref{fig:model7_agent_growth}. 
It can be observed that the increase in AI agent quantity significantly expands 
the scale of AI applications, ultimately contributing to overall economic output growth.

\subsection{AI Capability Level}

According to the \textit{Stanford AI Index Report 2025} \cite{StanfordHAI2025_TechPerformance}, 
the gap in artificial intelligence development between China and the United States 
is rapidly narrowing. 
On the LMSYS \textit{Chatbot Arena} platform, 
the performance advantage of the best U.S. model over the best Chinese model 
was 9.3\% in January 2024, 
but by February 2025, this gap had shrunk to only 1.7\%. 
In 2023, leading U.S. models maintained a considerable advantage 
over their primary Chinese counterparts, 
but by early 2025, this advantage had almost disappeared. 

Specifically, in authoritative benchmark evaluations such as 
MMLU, MMMU, MATH, and HumanEval, 
the performance differences between Chinese and U.S. models 
narrowed sharply from 17.5\%, 13.5\%, 24.3\%, and 31.6\% at the end of 2023 
to 0.3\%, 8.1\%, 1.6\%, and 3.7\% by the end of 2024. 
In the large language model competition landscape, 
the performance gap between Chinese and U.S. models 
has now fallen within a 30-point margin.

\begin{figure}[htbp]
    \centering
    \includegraphics[width=0.45\textwidth]{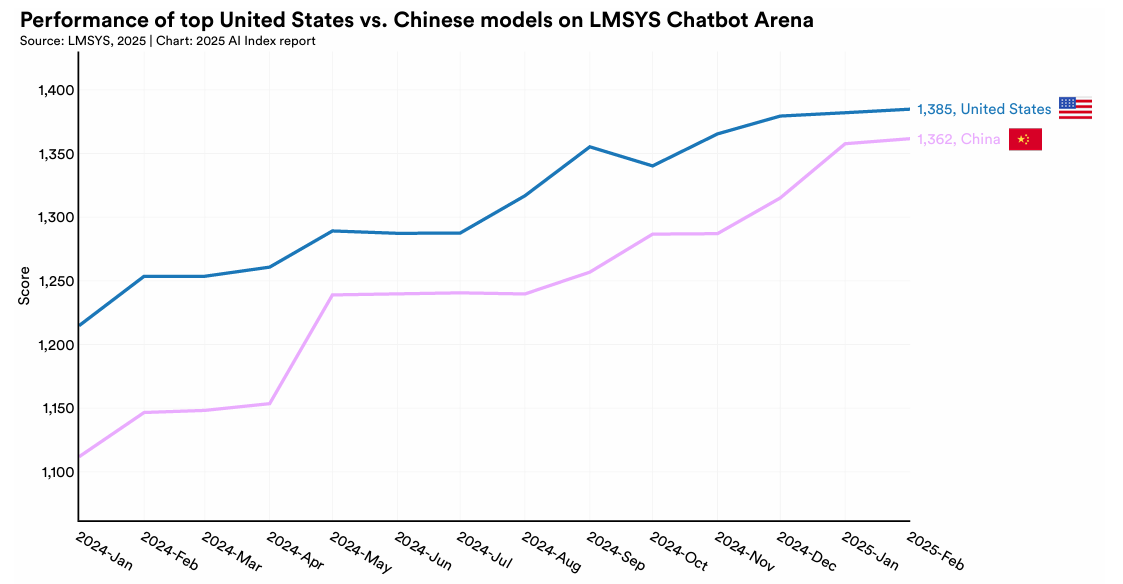}
    \caption{Performance comparison between U.S. and Chinese models 
    on the LMSYS Chatbot Arena platform}
    \label{fig:model8_arena_performance}
\end{figure}

To quantitatively represent China’s convergence trend in AI capability, 
this study models China’s AI efficiency parameter $\phi_{A,\text{cn}}(t)$ 
as a time-dependent function, 
while the U.S. AI efficiency parameter $\phi_{A,\text{us}}(t)$ 
remains constant at 688. 
The Chinese AI efficiency parameter is determined by 
the relative performance gap $\Delta(t)$ between the two countries:

\begin{equation}
\phi_{A,\text{cn}}(t) = 
\frac{\phi_{A,\text{us}}}{1 + \Delta(t)},
\end{equation}

where $\Delta(t)$ denotes the relative performance lead 
of U.S. models over Chinese models at time $t$ 
(measured in years since the beginning of 2019).

Based on benchmark results reported in the \textit{Stanford AI Index Report 2025}, 
the average performance gap between Chinese and U.S. models 
on the MMLU, MMMU, MATH, and HumanEval tests 
was 21.7\% at the end of 2023, 
corresponding to $\phi_{A,\text{cn}}(2023) = 565$; 
by the end of 2024, the gap narrowed to 3.4\%, 
yielding $\phi_{A,\text{cn}}(2024) = 665$. 
By fitting these data, 
the performance gap function can be expressed as:

\begin{equation}
\Delta(t) = \Delta_0 \exp\!\left(-\frac{t}{\tau}\right)^{\beta},
\quad \text{where } \tau = 4.27, \ \beta = 5.89.
\end{equation}

\begin{figure}[htbp]
    \centering
    \includegraphics[width=0.45\textwidth]{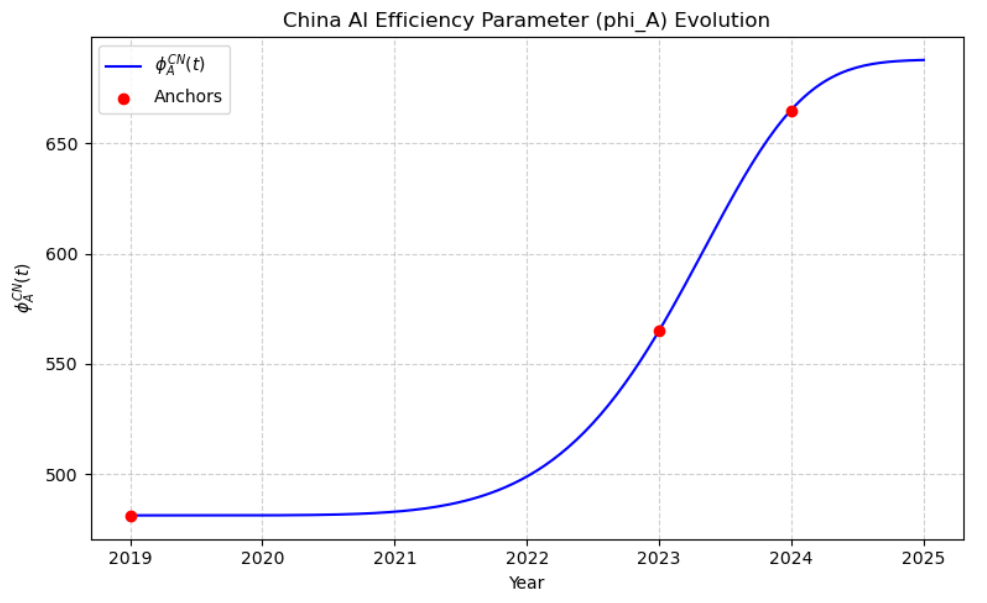}
    \caption{Fitted curve of China’s AI efficiency parameter $\phi_{A,\text{cn}}(t)$}
    \label{fig:model9_phiA_fit}
\end{figure}

\begin{figure}[htbp]
    \centering
    \includegraphics[width=0.45\textwidth]{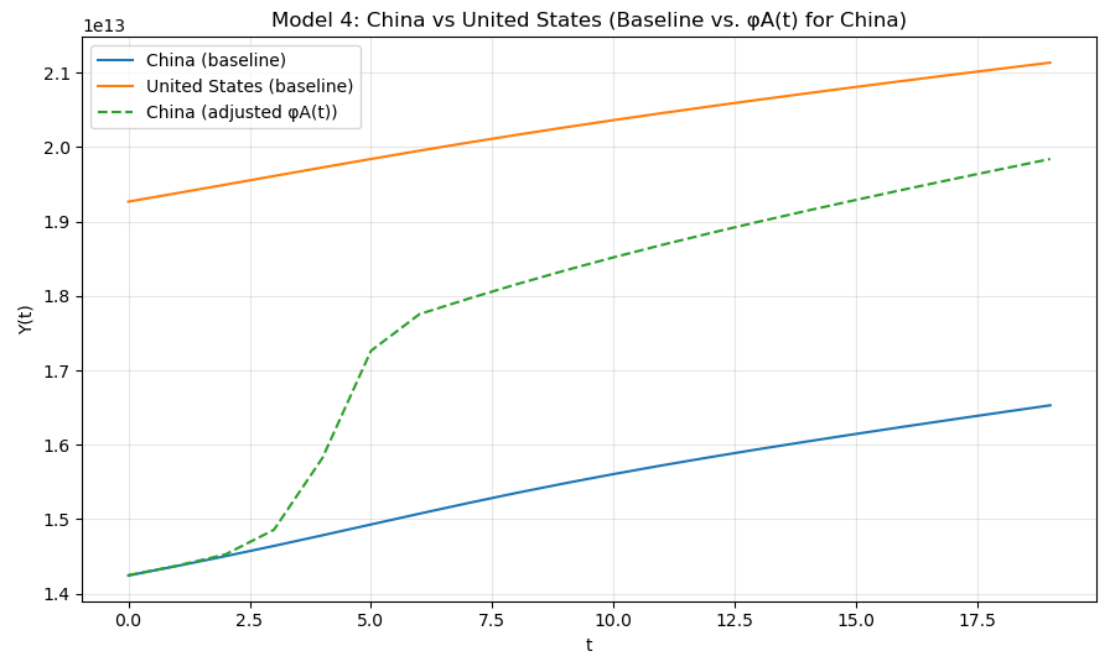}
    \caption{Comparison of total social output 
    after adjusting AI capability parameter $\phi_A$}
    \label{fig:model10_phiA_output}
\end{figure}

Under this parameter setting, 
the simulation results indicate that China’s total social output 
increases substantially, 
as shown in Figure~\ref{fig:model10_phiA_output}. 
The improvement reflects the compounded impact 
of AI capability enhancement and productivity diffusion over time.

\subsection{Comprehensive Adjustment}

The preceding analyses indicate that increasing the number (scale) of AI agents 
and enhancing the level (technology) of AI capability 
are the two key pathways driving output growth. 
This subsection integrates both dimensions to examine their combined effects. 
Simulation results show that when both the AI agent growth rate 
and the AI efficiency parameter are simultaneously adjusted, 
the resulting synergistic effect is substantially greater 
than in scenarios involving a single policy adjustment.

\begin{figure}[htbp]
    \centering
    \includegraphics[width=0.45\textwidth]{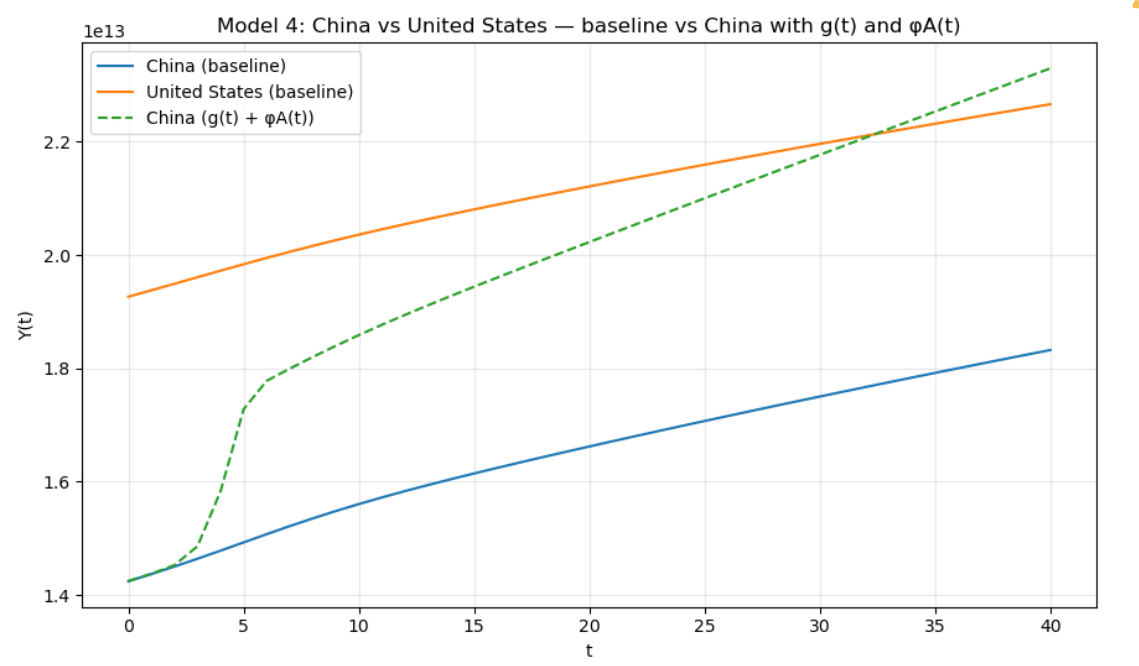}
    \caption{Comparison of total social output 
    after simultaneous adjustment of AI agent quantity and AI capability level}
    \label{fig:model11_joint_adjustment}
\end{figure}

As illustrated in Figure~\ref{fig:model11_joint_adjustment}, 
under this comprehensive scenario, 
China’s total social output increases far more significantly 
than in either of the single-adjustment cases—
namely, adjusting only the AI agent growth rate (Figure~\ref{fig:model7_agent_growth}) 
or only the AI efficiency parameter (Figure~\ref{fig:model10_phiA_output}). 
This confirms the existence of a \textit{resonance effect} 
between application scale and technological efficiency: 
a larger deployment scale of AI agents provides a broader environment 
for the application of advanced technologies, 
while higher technical efficiency enhances the marginal return 
of each AI agent input. 
Together, these two mechanisms reinforce each other, 
substantially accelerating China’s catch-up momentum 
and significantly reshaping the comparative output landscape 
between China and the United States.

\section{Conclusion}

Building upon a multilayered agent-based economic model developed in prior research, 
this paper systematically simulates and compares the dynamic impacts 
of artificial intelligence (AI) diffusion on the economic growth structures 
of China and the United States. 
From the perspectives of resource allocation, AI collaboration, network effects, 
and aggregate social output, 
our study reveals a general pattern of 
``U.S. leading and China accelerating.'' 
China’s catch-up potential is primarily rooted in its accelerating trends 
of AI agent expansion and capability evolution.

\subsection{Key Findings}

Model results demonstrate that the United States, 
leveraging its first-mover advantages in computing infrastructure, 
technological maturity, and enterprise-level ecosystems, 
maintains a stable lead in output 
when AI functions as a ``collaborative enhancement tool'' 
(Models~2 and~3). 
Its developmental trajectory is characterized by 
a technology-driven and efficiency-oriented pattern.

However, as AI evolves into an ``independent productive entity'' 
(Models~4 and~5), 
the dominant drivers of growth shift 
from technological efficiency 
to the number of AI agents and their capability level. 
Simulation results show that China possesses remarkable potential in both aspects: 
its AI agent population continues to expand rapidly, 
and its technological capability is improving swiftly. 
The combined effects of these two dimensions generate 
substantial scale and synergy gains. 
This implies that, in the medium to long term, 
China may gradually narrow its output gap with the United States—
and under certain conditions, even surpass it—
through a dual-path strategy of 
\textit{scale expansion} and \textit{capability catch-up}.

\subsection{Strategic Implications}

The findings of this study offer clear strategic implications. 
For the United States, 
sustaining its efficiency advantage 
in core technologies is essential for maintaining its lead. 
For China, the strategic priority should center on a dual-engine approach. 
On one hand, 
large-scale deployment of AI agents 
and their cross-industry integration 
should be leveraged to strengthen network effects 
and industrial connectivity. 
On the other hand, 
greater investment in research and development is needed 
to overcome algorithmic and computational bottlenecks, 
enabling a transition 
from \textit{application-driven} growth to \textit{technology-led} development. 
Through this dual-engine strategy, 
China can transform its breadth of application scenarios 
into endogenous growth depth 
and build a sustainable, intelligent economic system.

\subsection{Limitations and Future Directions}

The conclusions of this study are based on 
model design and parameter assumptions. 
Although historical data fitting is used to enhance reliability, 
real-world dynamics—such as technological breakthroughs, 
industrial policies, and geopolitical environments—
remain highly uncertain. 
Future research can be extended in several directions. 
First, refining the classification of AI agents 
to incorporate general-purpose foundation models, 
domain-specific AI, and embodied intelligence 
within a unified analytical framework. 
Second, integrating multi-country technology diffusion 
and policy response mechanisms 
to capture the global dynamics of AI-driven growth. 
Third, combining empirical data 
to validate the marginal contributions 
of different mechanisms within real economic systems, 
thereby constructing a more explanatory and predictive 
dynamic modeling framework.

\bibliographystyle{IEEEtran}
\bibliography{main}

\end{document}